\documentclass{INTERSPEECH2023}

\interspeechcameraready

\title{Modular Speech-to-Text Translation for Zero-Shot Cross-Modal Transfer}
\name{Paul-Ambroise Duquenne$^{1,2}$, Holger Schwenk$^1$, Benoît Sagot$^2$}
\address{
  $^1$Meta AI\\
  $^2$Inria}
\email{padqn@meta.com, schwenk@meta.com, benoit.sagot@inria.fr}

\begin{document}

\maketitle
 \begin{abstract}
Recent research has shown that independently trained encoders and decoders, combined through a shared fixed-size representation, can achieve competitive performance in speech-to-text translation. In this work, we show that this type of approach can be further improved with multilingual training. We observe significant improvements in zero-shot cross-modal speech translation, even outperforming a supervised approach based on XLSR for several languages.
\end{abstract}

%\noindent\textbf{Index Terms}: speech translation, machine translation, sentence embeddings

\section{Introduction}
Speech Translation (ST), also called speech-to-text (S2T) translation, is a growing subfield in machine learning research \cite{berard2016listen,bansal2017towards,weiss2017sequence}. Its goal is to directly translate speech in a source language into text in a target language. To bridge the gap with cascaded systems, which first transcribe and then translate the input, some work integrated Machine Translation (MT) data into the training \cite{dong2021listen,liu2020bridging,alinejad2020effectively,ye2022cross,zhang2022bridging}. 

Recently, \cite{escolano2021enabling} and \cite{tmodules} presented zero-shot cross-modal speech translation results, with speech encoders compatible with text decoders. In particular, \cite{tmodules} introduce \mbox{T-Modules}, an architecture based on monolingual encoders for speech and text for multiple languages, independently trained with a teacher-student approach to fit a shared fixed-size representation. Decoders are independently trained on top of this shared representation, and can be combined with any encoder at test time, enabling zero-shot cross-modal translation.

However, using monolingual encoders is a limitation of \mbox{T-Modules}, %\cite{tmodules},
since multilingual models are known to allow for cross-lingual transfer and often better performance. In this paper, we investigate how a modular architecture such as \mbox{T-Modules} can lead to better results when multilingual training is used. We successively investigate the impact of: 
\begin{itemize}
\item Multilingual text encoders
\item An English text decoder trained on multilingual embedding inputs
\item Multilingual speech encoders, either combining all languages at hand or only those in the same language family. 
\end{itemize}
We carry out experiments on German, French, Spanish, Catalan, Turkish, Japanese and Mongolian-to-English text-to-text and speech-to-text translation. We show that using a fixed-size representation bottleneck allows for straightforward cross-modal transfer, allowing for instance our speech encoder trained on Romance languages combined with our decoder trained on text data only to outperform XLSR supervised ST results on CoVoST2 \cite{covost2:2020:arxiv}.

\begin{figure}[t!]
    \centering
    \includegraphics[width=1\linewidth]{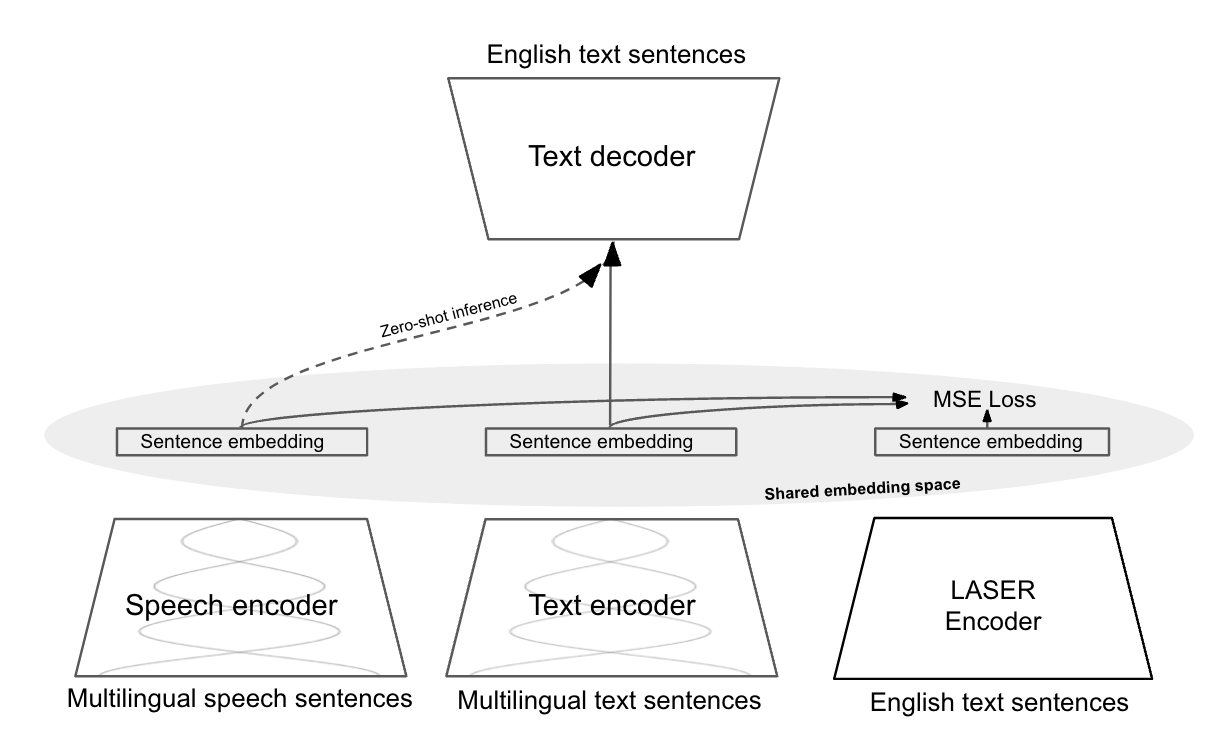}
    \caption{Overall architecture.}
    \label{fig:speech_student}
\end{figure}

\section{Related work} Leveraging MT data, in addition to ST data, for ST tasks is challenging due to the modality gap between speech and text representations. Several types of regularization, such as distance \cite{dong2021listen,liu2020bridging}, contrastive \cite{ye2022cross} and adversarial \cite{alinejad2020effectively,zhang2022bridging} regularization, have been proposed to address this issue.

In order to leverage unlabeled speech data, several pre-training methods have been introduced for multilingual speech \cite{babu2021xls,wav2vec2}. These pre-trained models were successfully used to train state-of-the-art ST models \cite{babu2021xls,li2020multilingual} and display interesting unsupervised capabilities on unseen languages thanks to cross-lingual transfer.
More recently, \cite{mslam} and then \cite{chen2022maestro} jointly pre-trained models on text and speech, before fine-tuning them on MT and ST data, reaching state-of-the-art ST results. With mSLAM \cite{mslam}, the authors explored zero-shot cross-modal transfer with its jointly trained encoder. It appears that cross-modal transfer works to some extent from speech to text but does not work at all from text to speech. \cite{escolano2021enabling} specifically explored zero-shot cross-modal transfer from text to speech. 

Recently, we saw the emergence of joint  fixed-size representation for speech and text for mining purposes \cite{Duquenne:2021:nips_mine,duquenne2022speechmatrix,khurana2022samu}: speech and text utterances are encoded in a shared sentence embedding space, then distances are computed in the embedding space to decide whether sentences should be considered parallel or not. Interestingly, using their \mbox{T-Modules} architecture, \cite{tmodules} explore how to decode such fixed-size representations into text or speech, reaching state-of-the-art performance in zero-shot ST and publishing the first zero-shot speech-to-speech translation results. 

\section{Methodology}

\begin{table*}
\centering\small
\begin{tabular}{l|l|lllllll}
\toprule
\textbf{Encoder}                     & \textbf{Decoder}                      & \textbf{de}   & \textbf{fr}   & \textbf{es}   & \textbf{ca}   & \textbf{ja}   & \textbf{tr}   & \textbf{mn}    \\
\midrule
Student multilingual        & \{en,de\}\_en                & 43.6 & 45.4 & 32.5 & 43.5 & 26.1 & 36.7 & 23.0  \\
Student multilingual        & \{en,de,fr,es,ca,ja,tr,mn\}\_en & 43.7 & 45.6 & 32.3 & 44.5 & 26.4 & 37.1 & 23.8  \\
\midrule
\multicolumn{9}{l}{\textbf{Baselines (previous works)}} \\
\midrule
T-Modules (Stud. monoling.) & \{en,de\}\_en                & 44.2 & 44.9 & 32.6 & 40.7 & 26.5 & 37.3 & 19.4  \\
M2M100~                     & M2M100~                      & 44.7 & 45.5 & 31.1 & 42.5 & 26.1 & 36.9 & 20.9  \\
Deepnet                     & Deepnet                      & 48.0 & 49.9 & 35.2 & 46.2 & 32.7 & 44.2 & 23.9 \\
\bottomrule
\end{tabular}
\caption{BLEU on FLORES devtest for text-to-text xx-en translation with a multilingual student. We compare our results to massively multilingual supervised models, M2M100 \cite{m2m100} and Deepnet \cite{wang2022deepnet}.}
\label{tab:t2t_en}
\end{table*}

\begin{table*}
\centering\small
\begin{tabular}{l|l|llll}
\toprule
\textbf{Encoder}              & \textbf{Decoder}                      & \textbf{pt}  & \textbf{it}  & \textbf{nl}  & \textbf{id}   \\
\midrule
Student de           & \{en,de\}\_en                & 35.2 & 22.7 & \textbf{24.2} & \textbf{20.3}  \\
Student ca           & \{en,de\}\_en                & 36.6 & 24.6 & 18.0 & 15.4  \\
Student es           & \{en,de\}\_en                & \textbf{40.9} & \textbf{24.9} & 19.4 & 16.0  \\
\midrule
Student multilingual & \{en,de\}\_en                & 42.0 & 28.2 & 25.1 & 22.0  \\
Student multilingual & \{en,de,fr,es,ca,ja,tr,mn\}\_en & \textbf{43.3} & \textbf{28.6} & \textbf{25.7} & \textbf{24.2} \\
\bottomrule
\end{tabular}
\caption{BLEU on FLORES devtest  for \textbf{unsupervised} text-to-text xx-en translation.}
\label{tab:t2t_zeroshot}
\end{table*}
\subsection{Original T-modules methodology}
In this paper, we base our work on \mbox{T-Modules} incremental learning procedure: for each language, text encoders are trained to encode text in a shared sentence embedding space, minimizing MSE loss to LASER English sentence embeddings. This approach, called teacher-student approach, requires xx-en bitext training data to train an xx student encoder. This training step forces all text translations to be encoded closely and is key for future efficient cross-lingual transfer.

As a second step, text decoders are trained for each language, learning to decode frozen sentence embeddings. Source sentences are encoded with previously trained text student encoders, whose weights are not updated during decoder training, and the trained decoder learns to decode target sentences with cross-entropy loss. The decoder is trained on both auto-encoding and translation tasks, requiring bitext training data.
For instance, to train a German text decoder, en-de bitexts are used to learn en-de translation task as well as de-de auto-encoding task.

Finally, for each language, speech encoders are trained to encode speech in this shared embedding space, minimizing MSE loss to their frozen transcription embeddings. Transcriptions are encoded with text encoders previously trained and are used as targets in this teacher-student training for speech.

This training procedure leads to sentence embeddings compatible between languages and modalities. At test time, each encoder can be combined with a text decoder, to perform zero-shot text-to-text translation or zero-shot speech-to-text translation. 
\begin{table*}[h]
\centering\small
\resizebox{\textwidth}{!}{%
\begin{tabular}{l|l|lllllll}
\toprule
\textbf{Encoder} & \textbf{Decoder} & \textbf{de} & \textbf{fr} & \textbf{es} & \textbf{ca} & \textbf{tr} & \textbf{ja} & \textbf{mn} \\
\midrule
\multicolumn{9}{l}{\textbf{Our models with no cross-modal supervision and full cross-lingual supervision}}\\
\midrule
Monoling. XLSR student (transcription)                                                  & \{en,de,fr,es,ca,ja,tr,mn\}\_en & 33.0         & 37.3         & 37.7         & 29.4         & 12.0         & 6.1          & 0.8          \\
\midrule
Multiling. XLSR student (transcription)                                                  & \{en,de,fr,es,ca,ja,tr,mn\}\_en & 30.9         & 35.3         & 37.5         & 31.9         & 16.5         & 0.9          & 0.7          \\
\midrule
Romance XLSR student (transcription) & \{en,de,fr,es,ca,ja,tr,mn\}\_en & --- & 37.0 & 38.4 & 33.2 & --- & --- & --- \\
\midrule
Romance XLSR student (transcr. + transl.) & \{en,de,fr,es,ca,ja,tr,mn\}\_en & --- & 38.3 & 39.7 & 34.8 & --- & --- & --- \\
\midrule
\multicolumn{9}{l}{\textbf{Previous work: model with no cross-modal supervision and only partial cross-lingual supervision}}  \\
\midrule
T-Modules monoling. (transcr.) (our reimpl.) & \{en,de\}\_en                & 33.0         & 35.7         & 36.3*         & 27.9*         & 11.2         & 6.1          & 1.0          \\
\midrule
\multicolumn{9}{l}{\textbf{Previous work: models with cross-modal and cross-lingual supervision}}  \\
\midrule
XLSR & finetuned mBART & 33.6  & 37.6  & 39.2  & 33.8  & 16.7  & 3.5  & 1.6   \\
mSLAM & mSLAM decoder & 35.9  & 39.0  & 41.0  & 35.4  & 24.2  & 3.3  & 0.8   \\
Whisper Large & Whisper Large  & 34.3  & 34.4  & 38.0  & 30.3  & 26.7  & 24.2 & 0.3   \\
\bottomrule
\end{tabular}
}
 \caption{BLEU on CovoST2 test set for speech-to-text xx-en translation using a decoder trained on text for several xx-en directions. We compare our results to supervised baselines XLSR \cite{babu2021xls}, mSLAM \cite{mslam} and Whisper Large \cite{radford2022robust}. Note that the latter was trained on significantly more speech data. (*) we trained monolingual students, note that Spanish and Catalan were learnt together in the original paper.}
    \label{tab:s2t}
\end{table*}
\subsection{Our methodology}
However, the \mbox{T-Modules} architecture grows linearly with the number of languages, as independent encoders and decoders are trained for each language. 

In this work, we explore multilingual training of encoders for text and speech, lowering the number of modules, and allowing for efficient cross-lingual transfer. While we did not explore multilingual decoders, we leveraged multilingual training data to learn an English text decoder with multiple xx-en translation directions and noticed a better robustness of the decoder for cross-modal translation. 

While our main focus is speech-to-text translation, the \mbox{T-Modules} incremental training procedure, allows us to first analyze the performance of text modules on text-to-text machine translation, before evaluating the architecture in the cross-modal setting with speech-to-text translation. The following section develops the experiments we conducted for these multilingual trainings and the analysis of the cross-lingual and cross-modal transfers.

\section{Experiments and results}
\subsection{Multilingual training for text}

We follow the training procedure introduced in \mbox{T-Modules}, but train a multilingual text student. We focus on the same set of languages, namely English (en), German (de), French (fr), Spanish (es), Catalan (ca), Japanese (ja), Turkish (tr) and Mongolian (mn). The multilingual text encoder is initialized with \texttt{XMLR Large} \cite{xlmr} and finetuned with bitexts from CCMatrix \cite{schwenk:2021:acl_ccmatrix} involving English and all languages listed in the introduction on Tesla V100 GPUs for a few days. As in \cite{tmodules}, we use additional bitext data mined with LASER3 \cite{heffernan2022bitext} for Mongolian. 

We train the multilingual text encoder to fit the English LASER space \cite{Artetxe:2019:tacl_massive_ml}, minimizing the MSE loss between the outputs of our trained encoder and the LASER embeddings of the corresponding English translations. This procedure is commonly called teacher-student approach. The fixed-size representation is obtained by max-pooling the outputs of the trained encoder. We will evaluate this new multilingual student encoder by combining it with an English decoder (named \{en,de\}\_en decoder) trained to generate English translations from English and German sentence embeddings only, but that can be used on other languages at test time, as introduced in \mbox{T-Modules}. We used the exact same setting for this training as \mbox{T-Modules} paper, using de-en bitexts from CCMatrix, to be able to compare our translation performance.

Based on this multilingual student encoder, we trained a new English decoder with bitexts involving all our languages of focus with English. Indeed, \cite{tmodules} showed that adding de-en bitexts to an English decoder training significantly improved the translation performance compared to a decoder trained only on monolingual English data. We extend this idea to a more multilingual setting and analyse the translation performance of this new decoder. For decoder training, we follow the same architecture as in \mbox{T-Modules}, using a 12-layer transformer decoder. We name this new decoder: \{en,de,fr,es,ca,ja,tr,mn\}\_en decoder.

We present the MT scores on the FLORES devtest set \cite{flores} in Table~\ref{tab:t2t_en}, obtained by our new multilingual encoder combined with one or the other of our two English decoders (\{en,de\}\_en and \{en,de,fr,es,ca,ja,tr,mn\}\_en), together with the scores obtained by three reference systems, including the original \mbox{T-Modules} architecture. First, decoding the sentence embeddings produced by our multilingual student with the \{en,de\}\_en decoder shows significant boosts in BLEU score for ca-en and mn-en translation tasks, with respectively $+2.8$ and $+3.6$ BLEU gains compared to \mbox{T-Modules} work. For other language directions, we notice a performance degradation of $-0.24$ BLEU on average. Using the \{en,de,fr,es,ca,ja,tr,mn\}\_en decoder instead significantly improves the translation results compared to decoding with the \{en,de\}\_en decoder, with an additional gain of $+ 1.0$ and $+0.8$ for Catalan and Mongolian respectively. All other directions are also improved except for es-en. We hypothesise that Catalan, which has much less training data compared to French or Spanish, may benefit from cross-lingual transfer from those languages, and that Mongolian, which is our lowest resource language, may benefit from the larger training data size to avoid overfitting.

In order to better analyse the cross-lingual transfer happening thanks to this multilingual training, we evaluate the translation of languages unseen during student and decoder training. Indeed, \texttt{XLMR Large} was pre-trained on a larger set of languages. Aligning a subset of languages with LASER English embeddings may transfer to other languages in an unsupervised way thanks to cross-lingual pre-trained representations. Therefore, we encode unseen languages with different student encoders to analyse how unsupervised cross-lingual transfer occurs. We chose 4 different languages, Portuguese and Italian, which are Romance languages close to French, Spanish and Catalan; but also Dutch, a Germanic language close to German; and Indonesian which is not similar to any language used for student and decoder training. We evaluate BLEU scores on the FLORES devtest set in Table~\ref{tab:t2t_zeroshot}, using either the \{en,de\}\_en or \{en,de,fr,es,ca,ja,tr,mn\}\_en decoder.

As expected, with monolingual student encoders, Portuguese and Italian are better translated  using the Catalan or Spanish student encoders, while Dutch is better translated using the German student encoder. This highlights the impact of language similarity between the training and unseen languages. Remarkably, Indonesian works  better with using the German student encoder than with the Spanish and Catalan ones. Also, Spanish student works better than the Catalan student on all unseen languages which may come from the fact that the Spanish student was trained on much more training data. Moreover, we notice that our new multilingual student encoder outperforms all monolingual encoders by a high margin, thanks to cross-lingual transfer and larger training data size. 

Finally, plugging our new \{en,de,fr,es,ca,ja,tr,mn\}\_en decoder further improves the results. This shows that multilingual training for text may help for translating low-resource and unseen languages in the \mbox{T-Modules} architecture.

\subsection{Multilingual training for speech}

\begin{table*}
\centering\small
\begin{tabular}{l|l|llll}
\toprule
\textbf{Encoder}                    & \textbf{Decoder}                      & \textbf{pt}  & \textbf{it}   & \textbf{nl}  & \textbf{id}   \\
\midrule
De transcription student     & \{en,de,fr,es,ca,ja,tr,mn\}\_en & 0.4 & 1.8  & \textbf{3.6} & 0.2  \\
Es transcription student     & \{en,de,fr,es,ca,ja,tr,mn\}\_en &   \textbf{7.3}   &   \textbf{10.3}    &  0.8    &    0.2   \\
Ca transcription student    & \{en,de,fr,es,ca,ja,tr,mn\}\_en &  3.8    &  7.1     &  0.3    &   0.2    \\
\midrule
Romance transcription student       & \{en,de,fr,es,ca,ja,tr,mn\}\_en & \textbf{8.9} & \textbf{13.7} & 1.2 & 0.2  \\
Multilingual transcription student~ & \{en,de,fr,es,ca,ja,tr,mn\}\_en & 7.3 & 13.4 & \textbf{5.1} & \textbf{0.4}  \\
\bottomrule
\end{tabular}
 \caption{BLEU on CovoST2 test set for \textbf{unsupervised} speech-to-text xx-en translation.}
    \label{tab:s2t_unsup}
\end{table*}

In \mbox{T-Modules}, the authors apply the same teacher-student approach to learn speech student encoders to either fit transcriptions or translations encoded with text encoders. We respectively call such approaches, transcription teacher training and translation teacher training. At test time, the speech students can be plugged with decoders trained on text in order to perform zero-shot cross-modal translation as the decoder has never seen speech embeddings during training. 

Using our new \{en,de,fr,es,ca,ja,tr,mn\}\_en decoder, we explore multilingual training of speech student encoders for either all languages or grouping languages by family, finetuning XLSR 2B. In our languages of focus, we analyse the Romance family composed of French, Spanish and Catalan. Based on best results with students trained with a transcription teacher, we train a speech student with both transcription and translation as teachers which has previously shown best results \cite{tmodules}. We present results in Table~\ref{tab:s2t}.

First, we notice a $+0.7$ BLEU gain in average when plugging our new \{en,de,fr,es,ca,ja,tr,mn\}\_en decoder compared to \mbox{T-Modules} work that used a \{en,de\}\_en decoder, showing the importance of a multilingual training of the decoder. Second, we notice a performance degradation when training a multilingual speech student on all languages, especially for Japanese with $-5.2$ BLEU. Only Catalan and Turkish benefit from multilingual training  with respectively a $+2.5$ and $+4.5$ BLEU improvement. We hypothesise that grouping all languages with different acoustic and morphological properties may cause interference, which could explain the performance degradation on other languages. 

In order to overcome this issue, we suggest grouping languages by families. For our study, this means grouping French, Spanish and Catalan into the Romance family and train a Romance multilingual speech student encoder on these languages only. Using this training procedure, we mainly notice gains in speech translation performance, with $+1.4$ BLEU gain in average. When adding written translations in the teacher-student training for Romance encoder, we notice significant improvements, even outperforming XLSR supervised results for these languages, while being fully cross-modal zero-shot ST (our \{en,de,fr,es,ca,ja,tr,mn\}\_en decoder has never seen speech embeddings during training). 

Then, we conduct the same analysis we have done for text on unseen languages for speech. Indeed, XLSR was pre-trained on a larger set of languages and we want to study the cross-lingual transfer that can occur when performing unsupervised speech translation on languages unseen during student training. 

Similarly to our findings on text, the German speech student encoder works better on Dutch, while Catalan, Spanish and Romance student encoders work better on Portuguese and Italian. The Spanish student encoder works better than the Catalan one on these languages, due to larger training data size. Indonesian is not working in this unsupervised setting, because it has no similarity with the languages used for training. Moreover, our findings regarding trained languages hold for unseen languages: the Romance encoder works better on Portuguese and Italian than the fully multilingual student and the Catalan or Spanish student encoders. However, not surprisingly, the fully multilingual student encoder works better for Dutch than the Romance encoder or the monolingual German one. This highlights even more that smart multilingual training for speech, grouping languages by family, yields to best results.

\section{Conclusion}

While \cite{tmodules} focused with their \mbox{T-Modules} architecture on zero-shot cross-lingual and cross-modal transfer, we focused on zero-shot cross-modal transfer only, which is currently a challenging part of Speech Translation and showed promising results in \mbox{T-Modules} architecture. Compared to previous work, we showed that a modular architecture can outperform strong supervised baselines while being zero-shot cross-modal. 

Here, we focused on a limited set of languages, but going multilingual significantly improved the results. We expect the same tendency when further increasing the number of languages and hope that our results will encourage the development of compatible representation for speech and text in massively multilingual settings. Interestingly, on the speech side, we showed that going fully multilingual hurts the translation performance. However, we found that a language-family-wide encoder produces the best results while being easily trainable in such a modular framework.

This work shows that multilingual training significantly boosts results for zero-shot cross-modal speech-to-text translation. But \cite{tmodules} also integrated speech decoders in the \mbox{T-Modules} architecture, an interesting future research direction would be to study if the conclusions drawn from multilingual training for text-to-text and speech-to-text translation hold for speech-to-speech translation.


\newpage
\bibliographystyle{IEEEtran}
\bibliography{anthology,custom}

\end{document}